\documentclass{article}
\usepackage{iclr2025_conference,times}

\usepackage{hyperref}
\usepackage{url}

\usepackage{booktabs}
\usepackage{amsfonts}
\usepackage{nicefrac}
\usepackage{microtype}
\usepackage{xcolor}

\usepackage{graphicx}
\usepackage{natbib}
\usepackage{doi}
\usepackage{bm}
\usepackage{amsmath}
\usepackage{amssymb}
\usepackage{multirow}
\usepackage{cleveref}
\usepackage{caption}
\usepackage{subcaption}
\usepackage{wrapfig}
\usepackage{colortbl}

\newcommand{\modi}[1]{\textcolor{black}{#1}}
\newcommand{\neu}[1]{\textcolor{black}{#1}}

\definecolor{darkturquoise}{HTML}{06C2AC}
\definecolor{plum}{HTML}{DDA0DD}
\definecolor{deeppink}{HTML}{FF1493}
\newcommand{\reg}[1]{\textcolor{black}{#1}}
\newcommand{\inv}[1]{\textcolor{black}{#1}}

\newcommand{\Real}{\mathbb{R}}
\newcommand{\R}{\bm{R}}
\newcommand{\Rp}{\bm{R'}}

\newcommand{\x}{\bm{x}}

\newcommand{\transp}{\mathsf{T}}

\newcommand{\Loss}{\mathcal{L}}

\title{Revisiting the Relation Between Robustness and Universality}

\author{Max Klabunde\textsuperscript{*} \\
	University of Passau\\
	Passau, Germany \\
	\texttt{max.klabunde@uni-passau.de} 
    \And
        Laura Caspari\thanks{Equal contribution.}\\
	University of Passau\\
	Passau, Germany \\
	\texttt{laura.caspari@uni-passau.de} \\
    \And
    Florian Lemmerich \\
	University of Passau\\
	Passau, Germany \\
	\texttt{florian.lemmerich@uni-passau.de}
}

\iclrfinalcopy
\begin{document}

\maketitle

\begin{abstract}
The \emph{modified universality hypothesis} proposed by Jones et al.~(2022) suggests that adversarially robust models trained for a given task are highly similar.
We revisit the hypothesis and test its generality.
While we verify Jones' main claim of high representational similarity in specific settings, results are not consistent across different datasets.
We also discover that predictive behavior does not converge with increasing robustness and thus is not universal.
We find that differing predictions originate in the classification layer, but show that more universal predictive behavior can be achieved with simple retraining of the classifiers.
Overall, our work points towards partial universality of neural networks in specific settings and away from notions of strict universality.

\end{abstract}

\section{Introduction}
The universality hypothesis \citep{olah_zoom_2020} suggests that all trained neural networks for a given task are highly similar.
If this hypothesis held generally, interpretability research would be simplified, as insights for a specific model could be more easily transferred to other models.
While the hypothesis is unlikely to hold in a strict sense \citep{li_convergent_2016,breiman_statistical_2001}, \citet{jones_if_2022} proposed and presented evidence for a modified universality hypothesis (MUH): adversarial robustness may function as a strong prior on neural networks such that adversarially robust models will learn similar representations ``regardless of exact training conditions (i.e., architecture, random initialization, learning parameters)''.
They showed empirically that robust CNNs trained on ImageNet \citep{imagenet21a} are highly similar in the used input features of the data and in the representations they produce, whereas standard models are not.
Thus, training a single robust model is sufficient to mimic the behavior of any other or in their words ''if you've trained one, you've trained them all''.
Hence, analyzing specific robust models could provide general insights into how neural networks function.

However, their work has three key limitations which motivate us to revisit the link between robustness and universality. 
First, the experiments were centered around representational similarity, while one of the direct and arguably practically most relevant ways to study model similarity is to compare their predictions.
Second, a key part of the evidence was gathered with Centered Kernel Alignment (CKA) \citep{kornblith_similarity_2019}, a method to measure similarity of representations, which adopts a specific perspective on neural network similarity and was recently shown to have multiple pitfalls \citep{cui_deconfounded_2022,davari2023reliability,nguyen_origins_2022}.
Numerous other similarity measures have been proposed \citep{klabunde_similarity_2023,sucholutsky_getting_2023}, which provide alternative views on neural network similarity, which also leads to substantial differences in similarity estimates \citep{resi2024,soni_conclusions_2024,bo2024evaluating}.
Third, experiments exclusively used ImageNet as input data, which leaves the role of data uncertain, e.g., whether results transfer to other vision datasets or out-of-distribution data.

\begin{figure}
    \centering
    \begin{subfigure}[b]{0.5\textwidth}
         \centering
         \includegraphics[width=\textwidth]{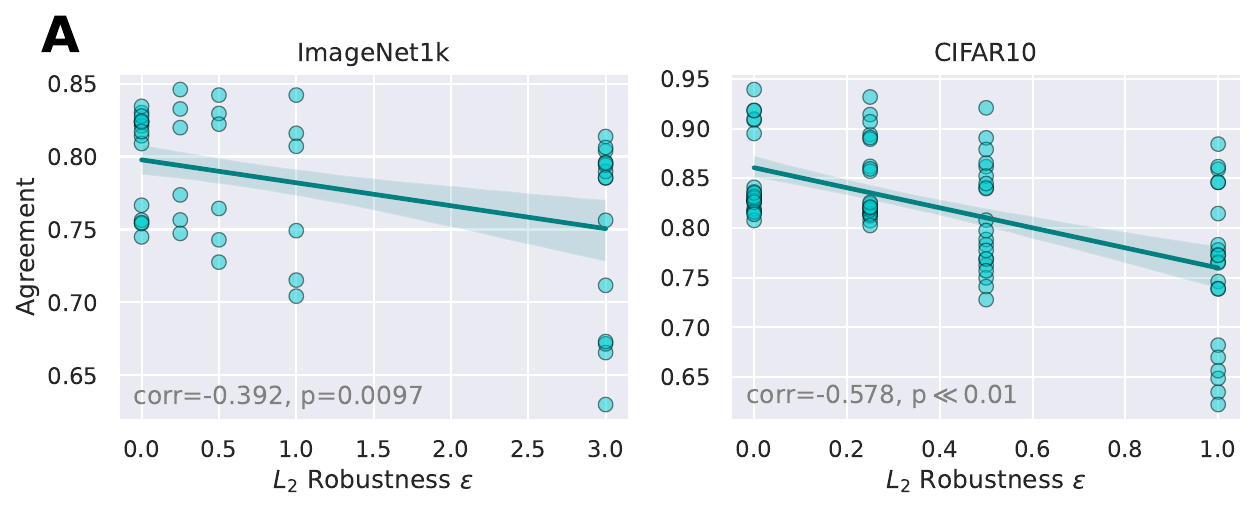}
     \end{subfigure}%
     \begin{subfigure}[b]{0.5\textwidth}
         \centering
         \includegraphics[width=\textwidth]{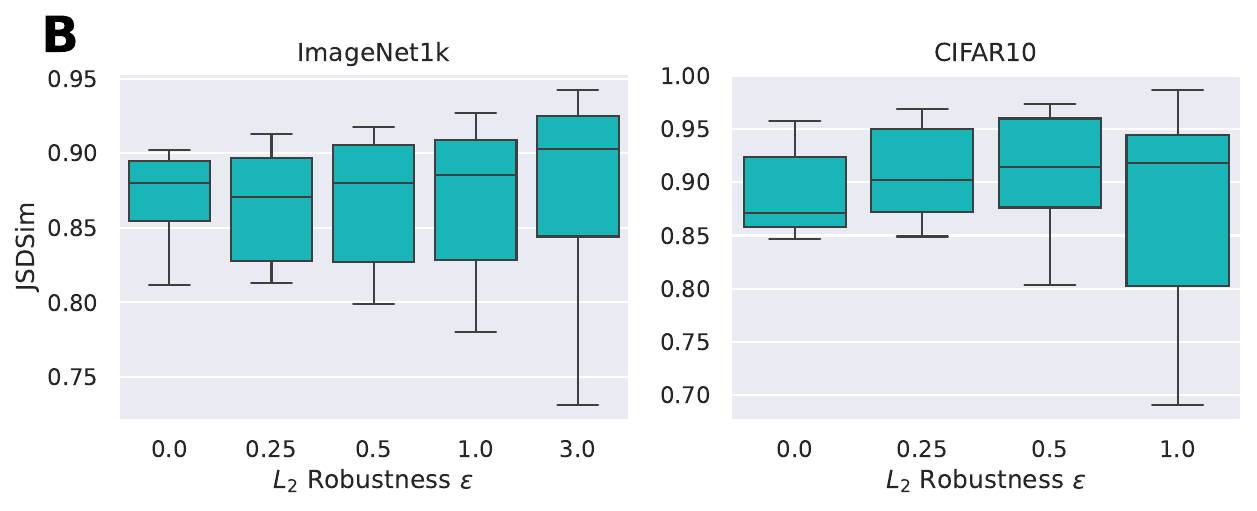}
     \end{subfigure}%
    \caption{\textbf{Predictive behavior remains distinct at high robustness contrary to the MUH.} 
    Distributions of agreement (\textbf{A}) and scaled Jensen-Shannon divergence (JSDSim) (\textbf{B}) across all model pairs when given \reg{regular images}.
    The MUH predicts that robust models converge to a universal solution, which should be reflected in highly similar predictions with increasing \neu{$L_2$} robustness $\epsilon$.
    However, predictions do not converge with increasing robustness, with agreement dropping and JSD showing increasing variance.
    This points towards an issue with the MUH.
    }
    \label{fig:predictions_regular}
\end{figure}

In this work, we thus critically reassess the modified universality hypothesis that suggests that all adversarially robust models for a given task are highly similar.
We conduct an extensive empirical study that involves multiple similarity measures, model architectures and datasets.
In contrast to previously published results, our study indicates that robust models should not be considered universal.
Our main contributions are:
\begin{enumerate}
    \item We show that predictions of robust models are not universal (see \Cref{fig:predictions_regular}).
    Their agreement scores do not converge with increasing robustness and the variance of Jensen-Shannon Divergence (JSD) scores increases with higher robustness levels (\Cref{subsec:preds_not_universal}).

    \item We verify that increasing robustness leads to more similar representations on ImageNet1k with a wider range of similarity measures. At the same time, some measures point towards lower absolute similarity than previously reported. Also, results are not robust to training dataset changes (\Cref{subsec:diffs_in_repmechs}).

    \item We identify that retraining classifiers on top of robust models can lead to higher predictive similarity and thus towards universality (\Cref{subsec:hypo2-classifier-diffs}).
\end{enumerate}

Code and data of our experiments are publicly available (see \Cref{apx:code_and_data}).

\section{Background and Methods}
\label{sec:methods}

\paragraph{Adversarial Robustness}
While neural networks achieve high performance in many tasks, they are susceptible to ---often imperceptible--- modifications of inputs that lead to wrong predictions \citep{szegedy_intriguing_2014}.
These modifications $\delta$ are usually computed via a constrained optimization problem:
\begin{equation}
    \delta^* = \arg \max_{\delta} \Loss (f(x+\delta), y) \quad \text{s.t.} \quad \|\delta\|_p \leq \epsilon,
\end{equation}
where $\Loss$ is the loss function, $x, y$ the input and target, respectively, and $\epsilon$ is the strength of the adversarial attack, i.e., the maximal allowed modification of the input.
By augmenting training data with adversarial examples, the space of potentially good models is constrained and robust models are produced, which are less susceptible to such attacks \citep{madry_towards_2019}.
For these models, perturbations need to be larger to induce misclassifications.
The larger $\epsilon$ for the adversarial examples, the more robust the model will be, but usually at the cost of lower accuracy on regular data.

\paragraph{Comparing Predictive Behavior}
\label{subsec:funcsim}
A simple test for universality is comparing predictions of models.
If models are universal, we should expect highly similar predictions.
Hence, we compare the predicted probability distributions and classifications using \modi{Jensen-Shannon Divergence (JSD) averaged over inputs} and the agreement rate.

For JSD, we normalize the outputs of the last network layer with a softmax, then compute:
\begin{equation}
    \operatorname{JSD}(\bm{L}, \bm{L'}) = \frac{1}{N} \sum_{i=1}^N \frac{1}{2}\operatorname{KL}(\bm{L}_i \| \bar{\bm{L}}_i) + \frac{1}{2}\operatorname{KL}(\bm{L'}_i \| \bar{\bm{L}}_i),
\end{equation}
where $\bm{L}, \bm{L'} \in \Real^{N \times C}$ are the collections of the predicted class probabilities, i.e., the softmaxed logits, for $C$ classes and $N$ fixed inputs, \modi{$\bar{\bm{L}} = 0.5 \cdot (\bm{L} + \bm{L'})$,} and $\operatorname{KL}$ is the Kullback-Leibler Divergence.
In the rest of the paper, we report JSDSim, i.e., scaled and normalized JSD to the range of $[0,1]$, such that a score of 1 indicates identical predicted distributions.

The agreement rate is the rate of instances that are predicted as the same class. This can be notated as the argmax of the logits, with $\mathbf{1}[\cdot]$ as the indicator function:
\begin{equation}
    \operatorname{Agreement}(\bm{L}, \bm{L'}) = \frac{1}{N} \sum_{i=1}^N \mathbf{1}[\arg\max_j \bm{L}_{ij} = \arg\max_j \bm{L'}_{ij}].
\end{equation}

\paragraph{Comparing Representations}
\label{subsec:repsim}
Another approach at testing universality is comparing the internal representations, i.e., the activation of a layer for some input.
Again, if models are universal, we expect that their internal processes are highly similar, which should lead to similar representations.
To measure representational similarity, activations are collected for a set of inputs resulting in a matrix $\R \in \Real^{N \times D}$, where $N$ is the number of inputs and $D$ is the number of neurons in the layer.
A representational similarity measure typically takes two such matrices as input and produces a single number that quantifies the similarity of these matrices.
The matrices may come from different layers or models, but are based on the same set of inputs such that the rows between the matrices correspond.
The similarity score respects certain transformations between representations that would keep them equivalent, e.g., switching neuron order, which would result in a different order of the columns of the compared matrices.
For a detailed introduction, we refer to the survey by \citet{klabunde_similarity_2023}.

In this work, we use four similarity measures: linear CKA \citep{kornblith_similarity_2019}, Orthogonal Procrustes (Procrustes) \citep{ding_grounding_2021,williams_generalized_2022}, k-NN Jaccard Similarity (Jaccard), and Representation Topology Divergence (RTD) \citep{barannikov_rtd_2022}.
Intuitively, these measures summarize the similarity of representations across multiple different aspects, e.g., specific properties of their geometry or topology.
These measures have been empirically shown to give meaningful similarity assessments \citep{resi2024}, but highlight different discrepancies between representations.
Thus, employing a set of similarity measures enables a more multi-faceted comparison of representations.
At the same time, the measures we use consider the same representations equivalent, i.e., any representations that only differ in rotation, reflection, scale, and translation.
This means we should expect similar similarity scores when representations are close to equivalent.

Formally, linear CKA computes a similarity score between 0 and 1 given two centered representations $\R \in \Real^{N \times D}, \Rp \in \Real^{N \times D'}$, i.e., with zero mean columns, as follows:
\begin{equation}
    \operatorname{CKA}(\R, \Rp) = \frac{\|\Rp^\transp \R\|_F^2}{\|\R^\transp \R\|_F \|\Rp^\transp \Rp\|_F},
\end{equation}
where $\|\cdot\|_F$ is the Frobenius norm. 
Based on the overall feature correlations, CKA measures global representational similarity.

Procrustes is another measure with a global view on similarity, but is a proper metric in contrast to CKA.
Procrustes finds the optimal orthogonal alignment between two representation spaces:
\begin{equation}
    \operatorname{Procrustes}(\R, \Rp) = \min_{\bm{Q}} \|\R\bm{Q} - \Rp\|_F = (\|\R\|_F^2 + \|\Rp\|_F^2 - 2 \|\R^\transp\Rp\|_*)^{1/2},
\end{equation}
where $\|\cdot\|_*$ is the nuclear norm, i.e., the sum of the singular values.
As $\R, \Rp$ need to have equal dimension for Procrustes, we zero-pad the representation with lower dimension.
In addition to zero-centering the columns, we scale the representation matrix to unit norm.
With this, we report $\frac{2-\operatorname{Procrustes}}{2}$ as ProcrustesSim, which is scaled to $[0,1]$, where 1 indicates maximal similarity.

We use Jaccard for a view on representation similarity that focuses on the similarity of the nearest-neighbor representations instead of the whole representation space.
Thus, Jaccard is a more local similarity measure.
Formally, Jaccard is defined as the average intersection over union of the nearest neighbors in the representation spaces:
\begin{equation}
    \operatorname{Jaccard}(\R, \Rp) = \frac{1}{N}\sum_{i=1}^N \frac{|\mathcal{N}_i^k(\R) \cap \mathcal{N}_i^k(\Rp)|}{|\mathcal{N}_i^k(\R) \cup \mathcal{N}_i^k(\Rp)|},
\end{equation}
where $\mathcal{N}_i^k(\R)$ are the $k$ nearest neighbors of the representation of input $i$ in $\R$. 
We use $k=10$ and cosine similarity on the centered representations to find the nearest neighbors.

Finally, RTD compares the topology of the representations. 
On a high level, RTD computes the strength of the discrepancy for different topological features. 
Summing up the strengths for all topological discrepancies yields a single number describing the overall topological divergence.
Hence, the score indicates distance between representations.
To make interpretation more consistent with the previous similarity measures, we report negative RTD, such that larger values mean higher similarity.
RTD does not have a fixed scale, making it difficult to interpret absolute levels of similarity, but still allows to examine similarity trends over different robustness levels.

\paragraph{Detecting Differences in the Representation Mechanism with Image Inversion} \label{par:image_inversion}

\begin{figure}
    \centering
    \includegraphics[width=1\textwidth]{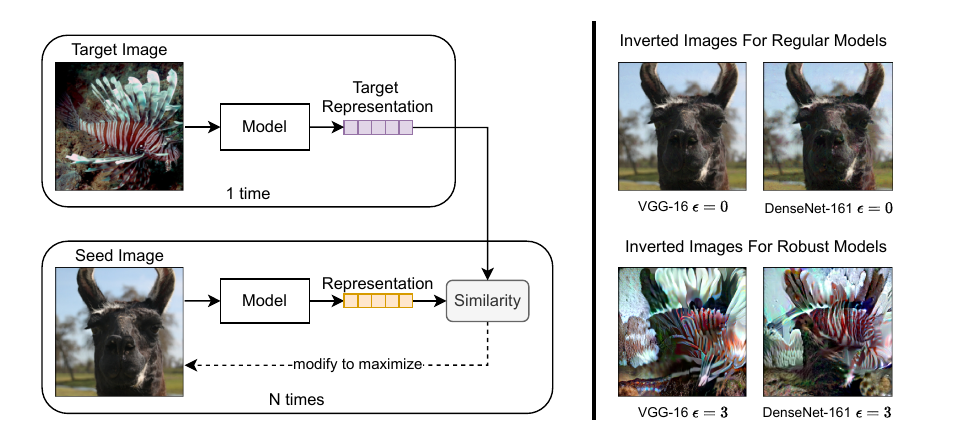}
    \caption{\textbf{Image inversion algorithm and examples of inverted images.}
    (Left) Inverted images are created by iteratively updating the seed image such that its representation becomes similar to that of the target image. It aims to introduce just the relevant features for the model and reduce feature cooccurrence.
    (Right) Examples of inverted images for VGG-16 and DenseNet-161 trained on ImageNet1k given the seed and target images shown on the left.
    The top row shows results for standard models ($\epsilon=0)$, the bottom one for robust models ($\epsilon=3$).
    The inverted images produced by robust and standard models are quite different.
    Inverted images of standard models are visually extremely similar to the seed image.
    For robust models, inverted images contain elements clearly belonging to the target image.
    They show how feature cooccurrence can be lessened, e.g., the dark background of the fish was not added to the image as both robust models mainly rely on the fins and texture of the fish.
    }
    \label{fig:inv_img_example}
\end{figure}

One problem of similarity measures is that they do not pick up on differences in the usage of input features as long as models produce similar representations or predictions \citep{jones_if_2022}.
This may lead to an overestimation of similarity between two neural networks.
We thus aim to test the similarity of the combination of the input feature reliance and the processing into a representation.
We call this combination the \emph{representation mechanism} and measure \emph{mechanistic similarity} \citep{lubana_mechanistic_2023}.

\emph{Image inversion} \citep{ilyas_adversarial_2019} presents a way to create a model-specific variant of an input that produces nearly the same representation as the original input, but consistently contains only those input features actually used by the model\footnote{\neu{The inverted images can be seen as model metamers \citep{feather_model_2023}.}}.
Hence, \inv{\emph{inverted images}} enable the study of the similarity of representation mechanisms.
If one model has a different mechanism that relies on another set of input features, it will not find those features in the inverted image of another model and thus will be unable to produce a similar representation.
Comparing the representations given inverted images gives us information about the similarity of the mechanisms.

To create an inverted image $\tilde{\x}$ for a given target image $\x$, a seed image $\bm{s}$ from a different class is modified such that it produces a representation similar to the representation of the target image.
More precisely, let $f^L(\x) \in \Real^D$ be the representation of model $f$ for the target image $\x$ in the penultimate layer $L$, then the inverted image $\tilde{\x}$ is computed as the output of
\begin{equation}
    \min_{\bm{s}} \frac{\|f^L(\bm{s}) - f^L(\x)\|_2}{\|f^L(\x)\|_2}.
\end{equation}
The optimization is done with gradient descent, so the naive solution of $s=x$ is not reached.
Instead, the most relevant input features for the model $f$ are introduced to the seed image.
As the seed image is sampled randomly from all images with a different class than the target image, feature cooccurrence in natural images, e.g., dog fur texture and dog ears, can be eliminated if only one of those features is relevant for $f$.
See \Cref{fig:inv_img_example} for an example.

\section{Experiments}
We will first lay out the general setup for the experiments to test the MUH.
As we will find surprising counter evidence to universality, we proceed by analyzing to what extent the MUH holds up.

\paragraph{Models} 
We use $L_2$-robust models trained on ImageNet1k \citep{imagenet21a}, ImageNet100 (a subset of ImageNet with 100 classes) and CIFAR-10 \citep{krizhevsky_learning_2009}.
The full list of models is given in \Cref{apx:models}.
While we train most of these models ourselves, we use the checkpoints released by \citet{salman_adversarially_2020} for ImageNet1k.
We study models with robustness of $\epsilon \in \{0, 0.25, 0.5, 1.0, 3.0\}$ on ImageNet1k and ImageNet100, but stop at $\epsilon=1$ for CIFAR-10 due to the lower resolution of images.

\paragraph{General Setup}
For each dataset mentioned above we compare the respective models using \reg{regular images} or \inv{inverted images} from the the dataset they were trained on as input.
For convenience, figures have color schemes corresponding to the type of input.
As inverted images are generated using a specific model, each pair of models A, B is compared twice, once on the inverted images generated by A and once on images generated by B.
All comparisons are made within one level of robustness and using the same dataset, i.e., A and B were always trained with the same $\epsilon$ and the same data.
To evaluate representational similarity with the measures outlined in \Cref{subsec:repsim}, we collect model activations at the penultimate layer.
For functional similarity, we apply a softmax to the model outputs to compute JSD and take the argmax of the logits as the predicted class for the agreement rate.
For a specific similarity measure, each comparison between a model pair A, B results in one similarity value.
Our analysis focuses on the distribution of these similarity values across all pairs.
\neu{The reported p-values are estimated with a permutation test.}

\subsection{Predictions of Robust Models Are Not Universal}
\label{subsec:preds_not_universal}

\begin{figure}
    \centering
    \begin{subfigure}[b]{0.5\textwidth}
         \centering
         \includegraphics[width=\textwidth]{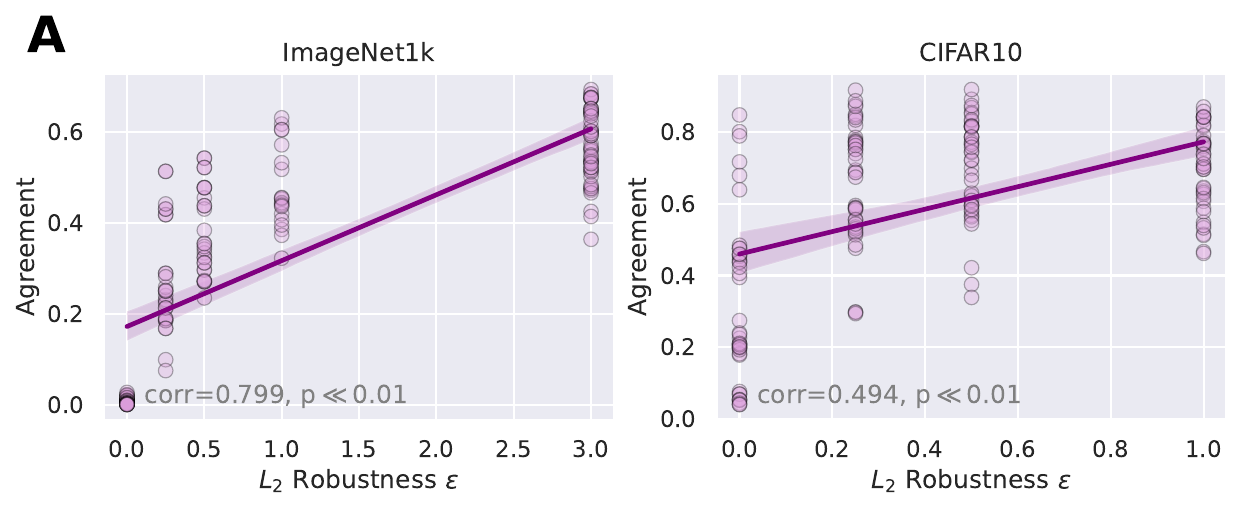}
     \end{subfigure}%
     \begin{subfigure}[b]{0.5\textwidth}
         \centering
         \includegraphics[width=\textwidth]{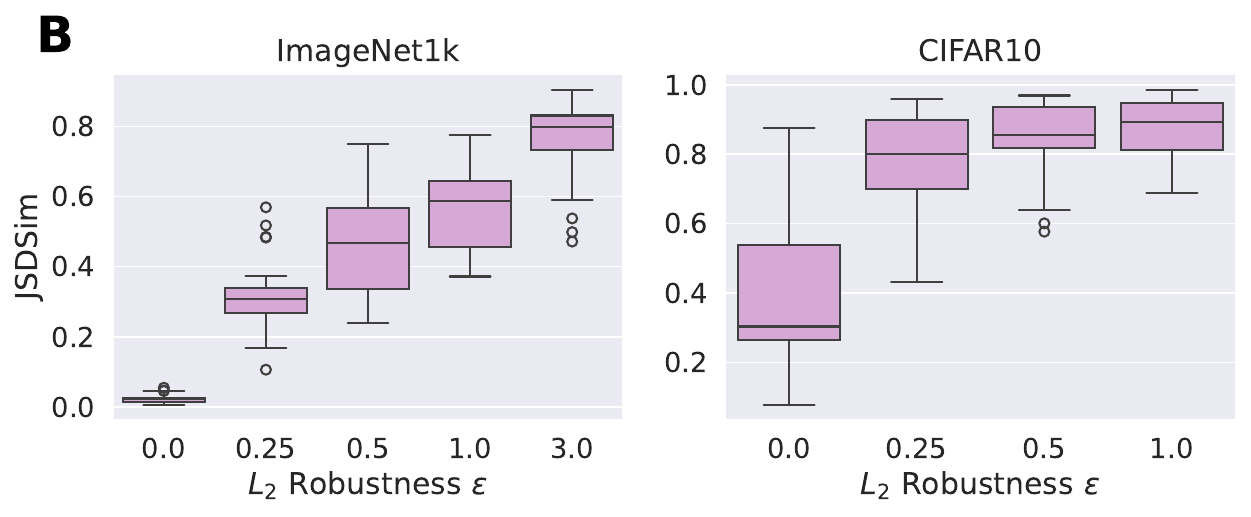}
     \end{subfigure}%
    \caption{\textbf{Similarity of predictions on \inv{inverted images} increases with robustness.} 
    Agreement (\textbf{A}) and JSDSim (\textbf{B}) distributions across all model pairs when given \inv{inverted images}.
    Both agreement and JSDSim increase with increasing robustness on both ImageNet1k and CIFAR-10.
    This means that robustness does lead to increased similarity in some aspects of the models, but arguably not to universality \modi{as the absolute similarity values still reveal differences between models.}
    }
    \label{fig:predictions_inverted}
\end{figure}

If adversarially robust models are universal in a strict sense, we would expect that their predictions overlap to a very high degree.
\Cref{fig:predictions_regular}A shows that this is not the case.
On \reg{regular images}, the agreement between predictions of highly robust models is much lower than the theoretical maximum agreement imposed by small accuracy differences \citep{fort2019deep}, \neu{see \Cref{apx:agreement_details}}. 
\modi{Instead, average agreement decreases with increasing robustness.}
Comparing the predicted distributions with Jensen-Shannon divergence (\Cref{fig:predictions_regular}B) instead of just the final predictions leads to the same conclusion: the predictive behavior is not universal.

However, using \inv{inverted images} as input, which highlights mechanistic similarity, reveals that robustness does have a profound impact on similarity of models.
\Cref{fig:predictions_inverted} shows how predictive behavior on these kind of inputs becomes more similar with increasing robustness.
\citet{jones_if_2022} showed similar effects for similarity of the representations.
Nevertheless, the differences in predictions given regular data must have an origin.
In the following sections, we will test multiple hypotheses and present a possible explanation.

\subsection{Hypothesis 1: Differing Predictions Stem From Differing Representations}
\label{subsec:diffs_in_repmechs}
Closely connected to the final predictions are the representations at the penultimate layer of the neural network as they are the input to the final classification layer.
Should these representations become more similar with increased robustness, we intuitively expect that the predictions also become more similar.
While dissimilar predictions are possible even if representations are similar, we expect this situation to be less likely.
Hence, as our first step, we inspect whether representations truly become more similar with increased robustness.

\subsubsection{Is Increased Similarity an Artifact of the Similarity Measure?}
\begin{figure}
    \centering
    \includegraphics[width=\linewidth]{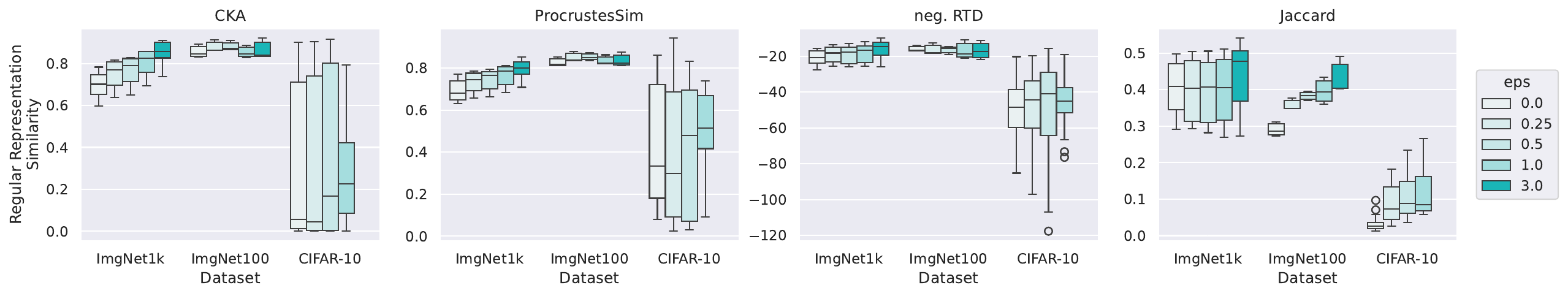}
    \includegraphics[width=\linewidth]{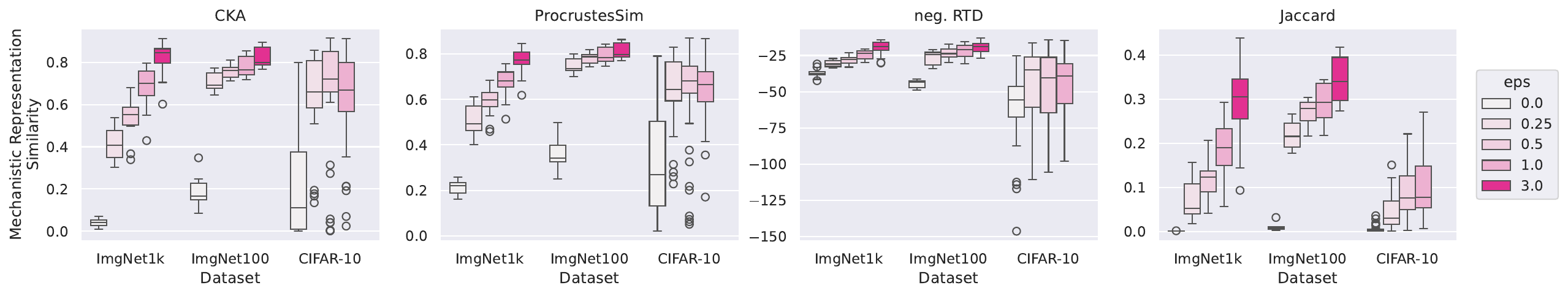}
    \caption{
        \textbf{Regular and mechanistic representational similarity over multiple similarity measures and datasets.}
        Contrary to recent work, multiple similarity measures generally agree on a positive correlation between robustness and similarity.
        However, results from ImageNet1k do not generalize to ImageNet100 and CIFAR-10 for regular similarity (top row), which lessens the generality of the MUH.
        For these two datasets, only Jaccard scores have significant Spearman rank correlation with robustness.
        At the same time, similarity is high on an absolute level for ImageNet100, but not for CIFAR-10.
        While results are consistent for mechanistic similarity (bottom row), robustness alone does not lead to universality.
    }
    \label{fig:repsim_distr_inverted}
\end{figure}

\citet{jones_if_2022} found high similarity between the representations of robust models using linear CKA, both in terms of regular similarity and mechanistic similarity.
Linear CKA is arguably the most popular similarity measure in the machine learning community, but was found to have several caveats \citep{cui_deconfounded_2022,davari2023reliability,nguyen_origins_2022}.
Additionally, recent work showed that results of representational similarity analysis are substantially influenced by the similarity measure \citep{resi2024,soni_conclusions_2024,bo2024evaluating}.
Thus, it is possible that representations do not become more similar in general with increased robustness, but only in the aspects that are measured by CKA---and these aspects could be of lesser importance with respect to influencing predictions.

We thus repeat the similarity measurements with three additional measures, namely ProcrustesSim, Jaccard, and negative RTD, as outlined in \Cref{sec:methods}. 
For Jaccard, we use a neighborhood size of 10 to encourage strict similarity assessment (see \Cref{apx:jaccard} for other neighborhood sizes).
\Cref{fig:repsim_distr_inverted} shows that similarity estimates are not majorly influenced by the similarity measure, given a fixed dataset.
\modi{While the results on ImageNet1k support the claims of MUH with respect to representational similarity}, we also repeat the experiments with models trained on different datasets.
In this case, we do not observe a clear relation between robustness and universality.
While mechanistic similarity remains significantly rank-correlated with robustness across all datasets, regular similarity with regular images as input does not follow the trend.
Instead, not only is the correlation insignificant, but the absolute similarity scores are also substantially different.

Overall, our results make it unlikely that the relation between robustness and representational similarity is dependent on CKA since other measures with different perspectives agree with the CKA results.
However, we identify that the  dataset is at least another factor influencing universality.
As this leads to some doubt of the validity of the MUH, we next investigate another possibility how representational similarity measures could give misleading similarity estimates. 
We will focus on models trained on ImageNet1k as they follow the MUH most closely so far.

\subsubsection{Does Unevenly Distributed Representational Similarity Explain Differing Predictions?}
\begin{figure}
    \centering
    \includegraphics[width=\textwidth]{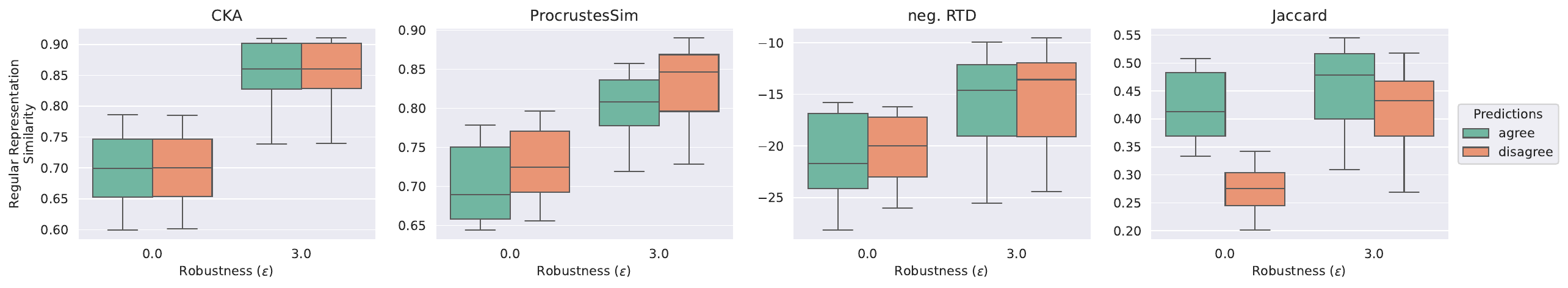}
    \caption{\textbf{Representational similarity of instances with agreeing or disagreeing predictions for ImageNet1k.}
    For the global representational similarity measures, representations of instances with disagreeing predictions are surprisingly more similar than those of agreeing predictions.
    Only the local Jaccard similarity ($k=10$) assigns slightly lower similarity to disagreeing instances.
    Hence, to explain the increasing disagreement with robustness, other parts of the models have to diverge as representations consistently become more similar with increased robustness.
    }
    \label{fig:repsim_subgroups}
\end{figure}

The previous experiment reported representational similarity over the full set of instances we use, condensed into a single number.
However, representational similarity for subsets of the data can be different.
For example, instances that are identically predicted by two models could have similar representations, whereas instances that are differently predicted have dissimilar representations. 
It is possible that information about such an uneven similarity distribution was lost in the aggregation over all instances.
If such an imbalance exists, it would be a simple explanation for the observed disagreement.
We thus compare the agreeing and disagreeing instances separately.
We focus on the most robust models with $\epsilon=3$ for ImageNet1k.

\Cref{fig:repsim_subgroups} shows that the similarity scores generally do not agree with this hypothesis.
Instead, representational similarity is even higher for instances with disagreeing predictions for the three global similarity measures.
With the local view of Jaccard similarity, similarity of disagreeing instances is lower, but only moderately.
The difference in medians corresponds to a difference of less than one neighbor, hence even locally representations are almost the same.

While our results show that similarity is indeed not homogeneous and subgroup-based analyses like proposed by \citet{kolling2023pointwise} could lead to better understanding of similarity, the differences in representational similarity are too small to explain the large differences in predictions of robust models.
Also, robust models are consistently representationally more similar than standard models.
Thus, the issue has to lie elsewhere---the parts of the models not analyzed yet are the classifiers.

\subsection{Hypothesis 2: Differing Predictions Originate in the Classifier}
\label{subsec:hypo2-classifier-diffs}

The previous results make it unlikely that the problem with the modified universality hypothesis originates in the representation extraction of the models.
We thus analyze the classifier as the remaining part of the model.

While representations are similar at the end of training, we do not know when they reach this state.
Prior work \citep{raghu_svcca_2017} showed models converge roughly bottom-to-top, i.e., first in the layers closest to the input and last in the final layers.
It is thus possible that the classifiers have "similar training data" only relatively late and thus are not able to converge to similar solutions in the remaining training steps.

\begin{wrapfigure}{r}{0.5\textwidth}
    \centering
    \includegraphics[width=0.45\textwidth]{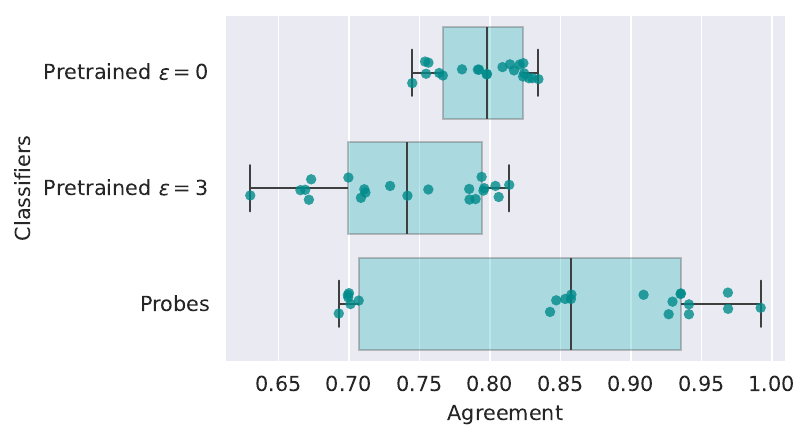}
    \caption{
        \textbf{Linear probes have higher agreement than pretrained classifiers.}
        While robust models ($\epsilon=3$) have lower agreement than standard models ($\epsilon=0$), training linear probes using robust representations as input enables more consistent predictions. 
        Probes consistently agree on a large share of the predictions even compared to standard models.
        The only exception is the cluster of low-agreement probe pairs that involve ResNet18 probes, which have low performance.
    }
    \label{fig:probes}
\end{wrapfigure}

To test whether it is possible to get more agreeing predictions based on the seemingly similar representations of robust models  with $\epsilon=3$, we remove the originally trained classifier and replace it with a freshly initialized linear layer, a probe.
We train this probe in a simple setup using the standard ImageNet1k training set over 30 epochs.
We use Adam with a learning rate of 0.005 and a cosine learning rate schedule.
We then compare the predictions of the probes using regular images (see \Cref{fig:probes}).

These probes have lower clean accuracy compared to the original models and lose some robustness. 
Typically, lower performance leads to lower agreement as more predictions can vary between any of the false classes.
However, in this case, the probes have higher agreement compared to the pretrained classifiers of both standard and robust models.
It is thus possible to influence the models towards universality, but the MUH would need to be further modified to take prediction agreement into account.

\section{Discussion} \label{sec:discussion}

\paragraph{Modified Universality Hypothesis Needs Another Modification}
We demonstrated that predictions of robust models do not converge with increasing robustness, which is in conflict to the MUH.
However, consistent with \citet{jones_if_2022}, we also observed that representations become more similar with increased robustness, from both a regular and a mechanistic perspective.
The contrast points towards an interesting direction for future work:
why is it that some aspects of models seem to be strongly constrained by robustness whereas others are not?
Also, how can unintuitive results be explained like higher representational similarity for disagreeing instances compared to similarity of agreeing instances?

\paragraph{Disconnection Between Behavioral and Representational Similarity}
Our findings indicate that relying solely on representational similarity scores can lead to misleading conclusions, as these scores can be disconnected from behavioral similarity.
While mechanistic representational similarity measures consistently increase with robustness, prediction agreement decreases.
Furthermore, most representational similarity measures do not show substantial differences between agreeing and disagreeing instances.
% We show that differing predictions originate at the classification layer, possibly because robust classifiers have not converged.
We thus argue that representational similarity measures should be viewed as exploratory tools rather than definitive indicators of model similarity. 
Any insights derived from these measures should be validated through additional experiments.
To be able to rely more on representational similarity measures, we believe that better theory and justification of similarity measures is necessary.
In the absence of such understanding, using multiple similarity measures could make findings slightly more robust.

\paragraph{Robust Models for Interpretability Research}
The ideal subject for interpretability research would give insights about many more models than just the model under study.
On the one hand, robust models likely partially fulfill this criterion--studying the representation mechanism could transfer across other robust models.
Models could also be modified towards universality as shown in \Cref{subsec:hypo2-classifier-diffs}.
On the other hand, our work is another point of evidence against universality in a strong sense, where all parts of a model are highly similar, and towards a world where models consist of universal and non-universal parts.
Studying universal parts may be of general interest, whereas non-universal parts may be only interesting for frontier models or specific models with high interest.
Hence, identifying universal components is an interesting direction of future work.

\paragraph{Increasing Robustness Beyond Our Experiments}
We observed that increasing robustness up to $\epsilon=3$ for ImageNet models leads to increased representational similarity of the models, e.g., the trend for CKA similarity appears to continue further.
Thus, extremely robust models may be a way to studying the whole model class at once--at least with respect to the aspects that make them similar from the CKA perspective.
However, increasing robustness even further would likely lead to further accuracy degradation.
Ultimately, such models may not be comparable to more widely used models, which could make detailed study of these models not worth it despite the aforementioned benefit.

\section{Related Work}
\label{sec:related_work}

\paragraph{Universality}
The question to what extent models are universal has attracted significant interest in prior work.
On the one hand, model multiplicity, i.e., the existence of multiple models with almost equal performance but different input-output behavior or representations, has been studied extensively \citep{breiman_statistical_2001,black_model_2022,heljakka_disentangling_2023}.
Architecturally similar models trained or updated on near-identical data can differ significantly \citep{klabunde_prediction_2023, somepalli_can_2022, marx_predictive_2020,black_leave-one-out_2021,liu_model_2022,mccoy_berts_2020,li_convergent_2016}.
Modifications to training or inference may be necessary to enforce consistent behavior between different models \citep{milani_fard_launch_2016,summers_nondeterminism_2021}.
In mechanistic interpretability, a more fine-grained view on universality is taken, i.e., whether the input-output behavior of a network is also implemented in the same way.
It leads to further evidence against universality \citep{zhong_clock_2023,chughtai_toy_2023}.

On the other hand, there is evidence for universality in certain scenarios.
Some features consistently appear in CNNs \citep{schubert_high-low_2021}.
Further, attention heads with specific functionality can be found across many transformer-based language models \citep{olsson_-context_2022,gould_successor_2023}.
Additionally, some of their internal processes for tasks such as indirect object identification \citep{merullo_circuit_2023} and retrieval \citep{variengien_look_2023} seem to be universal, at least across certain model classes.
On the smallest scale, certain neurons appear universal \citep{gurnee_universal_2024}.
Further, parts of two different models (trained for the same task) can be connected using simple transformations with little accuracy loss \citep{csiszarik_similarity_2021,bansal_revisiting_2021,lahner_direct_2023,moschella_relative_2023} indicating representational compatibility \citep{brown_understanding_2023}.
\neu{Models of the same architecture can recognize metamers generated for others at early layers \citep{feather_model_2023}.
This transfer ability improves if models use adversarial training.
Finally, \cite{huh_position_2024} found that model representations are converging, especially when model size increases or models are trained on multiple tasks, and posited the platonic representation hypothesis.}

While the two above collections of evidence for and against universality might seem contradicting, the scope of universality as well as what would be considered equivalent between networks differs drastically.
In fact, universality has multiple non-binary facets \citep{gurnee_universal_2024}.
Furthermore, universality may only occur for certain types of models \citep{jones_if_2022}.

\paragraph{Neural Network Similarity}
To measure similarity of neural networks, especially of their representations, numerous similarity measures have been proposed across machine learning and neuroscience \citep{klabunde_similarity_2023,sucholutsky_getting_2023}.
These measures represent different views on what kind of behavior is considered equivalent.
Due to its popularity, Centered Kernel Alignment (CKA) \citep{kornblith_similarity_2019} has attracted particular interest and was also used by \citet{jones_if_2022} who propose the hypothesis of universality across robust models.
However, several caveats of CKA are known: few data points may dominate the similarity score \citep{nguyen_origins_2022}, the choice of inputs may determine similarity measurements in early layers \citep{cui_deconfounded_2022}, and scores are generally brittle \citep{davari_inadequacy_2022}.

\section{Conclusion}
We revisit the modified universality hypothesis which states that adversarially trained models are highly similar.
We show that predictions of robust models are not universal as their agreement on regular images decreases with robustness.
While we further show that the representation mechanisms consistently become more similar with increased robustness, regular representational similarity does not consistently increase.
We demonstrate that these seemingly contradictory findings are likely the result of insufficient convergence at the classification layers.
More broadly, our analysis reveals that relying solely on representational similarity measures can be misleading as they do not capture relevant differences in models that lead to different predictive behavior.
Our results show that the modified universality hypothesis is not applicable to all components of robust neural networks.

\section*{Reproducibility Statement}
All code and data to reproduce our results are publicly available, see \Cref{apx:code_and_data} for details.

\bibliographystyle{unsrtnat}
\bibliography{references}

\newpage
\appendix

\section{Additional Model Information} \label{apx:models}
\subsection{ImageNet1k Models}
\label{apx:imagenet_models}

\begin{table}[t]
	\centering
 \caption{The number of parameters, accuracy (Acc) and adversarial accuracy (Adv. Acc.) for models trained on ImageNet1k. The adversarial accuracy of models with $\epsilon=0$ was evaluated with $\epsilon=0.25$. For the models marked in gray, we used the checkpoints provided by \cite{salman_adversarially_2020}.}
	\label{tab:img_cnns}
    \resizebox{\linewidth}{!}{
    \begin{tabular}{lccc|cc|cc|cc|cc}
		\toprule
		\multirow{2}{*}{\parbox{0.18\textwidth}{Architectures}} & \multicolumn{1}{c}{\multirow{2}{*}{Parameters}} & \multicolumn{2}{c}{$\epsilon = 0$} & \multicolumn{2}{c}{$\epsilon = 0.25$} & \multicolumn{2}{c}{$\epsilon = 0.5$} & \multicolumn{2}{c}{$\epsilon = 1$} & \multicolumn{2}{c}{$\epsilon = 3$} \\
		& & Acc. & Adv. Acc. & Acc. & Adv. Acc. & Acc. & Adv. Acc. & Acc. & Adv. Acc. & Acc. & Adv. Acc. \\
		\midrule
		ResNet-18 & 11.7M & \cellcolor{gray!25}69.80 & \cellcolor{gray!25}20.30 & \cellcolor{gray!25}67.42 & \cellcolor{gray!25}60.02 & \cellcolor{gray!25}65.48 & \cellcolor{gray!25}55.97 & \cellcolor{gray!25}62.31 & \cellcolor{gray!25}55.65 & \cellcolor{gray!25}53.11 & \cellcolor{gray!25}49.70 \\
		ResNet-50 & 25.6M & \cellcolor{gray!25}75.80 & \cellcolor{gray!25}25.97 & \cellcolor{gray!25}74.13 & \cellcolor{gray!25}67.42 & \cellcolor{gray!25}73.17 & \cellcolor{gray!25}64.23 & \cellcolor{gray!25}70.42 & \cellcolor{gray!25}64.32 & \cellcolor{gray!25}62.83 & \cellcolor{gray!25}59.47 \\
		Wide ResNet-50-2 & 68.9M & \cellcolor{gray!25}76.98 & \cellcolor{gray!25}29.37 & \cellcolor{gray!25}76.22 & \cellcolor{gray!25}69.82 & \cellcolor{gray!25}75.11 & \cellcolor{gray!25}66.70 & \cellcolor{gray!25}73.42 & \cellcolor{gray!25}67.36 & \cellcolor{gray!25}66.90 & \cellcolor{gray!25}63.45 \\
		Wide ResNet-50-4 & 223.4M & \cellcolor{gray!25}77.91 & \cellcolor{gray!25}32.74 & \cellcolor{gray!25}77.10 & \cellcolor{gray!25}72.82 & \cellcolor{gray!25}76.52 & \cellcolor{gray!25}69.00 & \cellcolor{gray!25}75.51 & \cellcolor{gray!25}62.78 & \cellcolor{gray!25}69.67 & \cellcolor{gray!25}45.17 \\
		ResNeXt-50 32x4d & 28.7M & \cellcolor{gray!25}77.32 & \cellcolor{gray!25}26.00 & - & - & 59.74 & 49.73 & 72.45 & 66.71 & \cellcolor{gray!25}65.92 & \cellcolor{gray!25}62.39 \\
		Densenet-161 & 25.0M & \cellcolor{gray!25}77.38 & \cellcolor{gray!25}28.78 & - & - & - & - & 60.12 & 13.33 & \cellcolor{gray!25}66.12 & \cellcolor{gray!25}62.72 \\
		VGG-16-BN & 138.4M & \cellcolor{gray!25}73.67 & \cellcolor{gray!25}10.86 & 68.49 & 61.57 & 68.32 & 59.29 & 66.33 & 60.14 & \cellcolor{gray!25}56.79 & \cellcolor{gray!25}53.51 \\
        % TinyViT & 5M & 72.65 & 24.03 & 71.16 & 65.27 & 69.30 & 60.87 & 66.49 & 60.58 & 56.45 & 53.22 \\
		\bottomrule
	\end{tabular}
    }
	
\end{table}
Table \ref{tab:img_cnns} shows all model architectures with their accuracy and number of parameters.
We use seven $L_2$-robust CNNs: ResNet-18, ResNet-50 \citep{he15a}, Wide ResNet-50-2, Wide ResNet-50-4 \citep{zagoruyko17a}, ResNeXt-50 32x4d \citep{xie17a}, Densenet-161 \citep{huang18a}, and VGG-16-BN \citep{simonyan15a}.

\paragraph{Training Details} \citep{salman_adversarially_2020} trained their $L_2$-robust ImageNet models for 90 epochs using an initial learning rate of $0.1$ which is reduced every 30 epochs by a factor of 10.
The training uses stochastic gradient descent (SGD) with a batch size of $512$, a momentum of $0.9$ and weight decay of $1e^{-4}$.
For standard training, cross-entropy was used as a loss function.
Robust training was conducted using projected gradient descent (PGD) \citep{madry_towards_2019} allowing $L_2$ perturbation of the respective $\epsilon$ value.
Adversarial examples were generate in three attack steps with a step size of $\frac{2}{3}\epsilon$.
We used an identical setting for training the remaining ImageNet1k models.

\paragraph{Inverted Images}
Inverted images were generated on the ImageNet Large Scale Visual Recognition Challenge 2012 (ILSVRC2012) validation set using the Robustness library \citep{engstrom19a}.
First 10,000 target images were randomly sampled from the dataset.
Then, for each sampled image, a seed image was sampled at random.
If the sampled seed image had the same class as the target, a new image was sampled until seed and target classes were different.
To generate an inverted image, the seed image was modified in three steps and the best result taken.

\subsection{ImageNet100 Models}
\label{apx:imagenet100_models}
\begin{table}[b]
	\centering
 \caption{The number of parameters, accuracy (Acc) and adversarial accuracy (Adv. Acc.) for models trained on ImageNet100. The adversarial accuracy of models with $\epsilon=0$ was evaluated with $\epsilon=0.25$.}
	\label{tab:img100_cnns}
    \resizebox{\linewidth}{!}{
	\begin{tabular}{lccc|cc|cc|cc|cc}
		\toprule
		\multirow{2}{*}{\parbox{0.18\textwidth}{Architectures}} & \multicolumn{1}{c}{\multirow{2}{*}{Parameters}} & \multicolumn{2}{c}{$\epsilon = 0$} & \multicolumn{2}{c}{$\epsilon = 0.25$} & \multicolumn{2}{c}{$\epsilon = 0.5$} & \multicolumn{2}{c}{$\epsilon = 1$} & \multicolumn{2}{c}{$\epsilon = 3$} \\
		& & Acc. & Adv. Acc. & Acc. & Adv. Acc. & Acc. & Adv. Acc. & Acc. & Adv. Acc. & Acc. & Adv. Acc. \\
		\midrule
		ResNet-50 & 25.6M & 79.00 & 45.92 & 79.28 & 73.60 & 77.16 & 68.44 & 74.30 & 60.86 & 69.88 & 47.54 \\
		Wide ResNet-50-2 & 68.9M & 80.50 & 51.44 & 80.22 & 74.88 & 79.92 & 72.40 & 75.64 & 63.98 & 69.22 & 46.56 \\
		Densenet-161 & 25.0M & 83.30 & 57.48 & 83.24 & 78.32 & 82.16 & 75.22 & 81.20 & 69.60 & 76.26 & 54.24 \\
		VGG-16-BN & 138.4M & 82.52 & 41.36 & 80.02 & 74.56 & 78.62 & 69.46 & 73.86 & 62.06 & 66.46 & 45.76 \\
		\bottomrule
	\end{tabular}
    }
	
\end{table}

Table \ref{tab:img100_cnns} shows accuracy scores for ImageNet100 models.

\paragraph{Training Details}
We trained the ImageNet100 models using the same training procedure as for ImageNet.

\paragraph{Inverted Images}
The process for generating inverted images is identical to that on ImageNet. Seed and target images were sampled from the ImageNet100 train set.

\subsection{CIFAR-10 Models}
\label{apx:cifar10_models}
Table \ref{tab:cif_cnns} shows the accuracy and number of parameters of each CIFAR-10 CNN.

\paragraph{Training Details} The CIFAR-10 models were trained using almost the same configuration as the $L_2$-robust ImageNet1k CNNs.
The only modification for standard training was using a weight decay of $5e^{-4}$.

\paragraph{Inverted Images}
Seed and target images were taken from the CIFAR-10 test set, which contains 10,000 images.

\begin{table}[b!]
	\centering
        \caption{The number of parameters, accuracy (Acc.) and adversarial accuracy (Adv. Acc.) for models trained on CIFAR-10. The adversarial accuracy of models with $\epsilon=0$ was evaluated with $\epsilon=0.25$.}
	\label{tab:cif_cnns}
    \resizebox{\linewidth}{!}{
	\begin{tabular}{lccc|cc|cc|cc}
		\toprule
		\multirow{2}{*}{\parbox{0.18\textwidth}{Architectures}} & \multicolumn{1}{c}{\multirow{2}{*}{Parameters}} & \multicolumn{2}{c}{$\epsilon = 0$} & \multicolumn{2}{c}{$\epsilon = 0.25$} & \multicolumn{2}{c}{$\epsilon = 0.5$} & \multicolumn{2}{c}{$\epsilon = 1$} \\
		& & Acc. & Adv. Acc. & Acc. & Adv. Acc. & Acc. & Adv. Acc. & Acc. & Adv. Acc. \\
		\midrule
		ResNet-18 & 11.2M & 93.20 & 2.84 & 81.86 & 49.02 & 90.30 & 26.96 & 86.14 & 41.99 \\
		ResNet-50 & 23.5M & 90.03 & 0.17 & 86.10 & 25.52 & 78.73 & 37.67 & 71.34 & 45.59 \\
		Wide ResNet-50-2 & 66.9M & 83.31 & 1.59 & 77.81 & 23.28 & 71.99 & 33.13 & 60.05 & 38.53 \\
		Wide ResNet-50-4 & 221.4M & 82.52 & 1.67 & 78.47 & 22.88 & 70.09 & 31.81 & 59.21 & 38.60 \\
		ResNeXt-50 32x4d & 26.5M & 81.45 & 2.06 & 77.53 & 22.52 & 68.17 & 32.12 & 57.48 & 36.89 \\
		Densenet-161 & 23.0M & 94.22 & 1.00 & 91.91 & 29.14 & 88.27 & 43.99 & 83.60 & 48.18 \\
		VGG-16 & 14.7M & 91.20 & 0.02 & 87.88 & 22.72 & 82.38 & 37.67 & 70.20 & 46.63 \\
		\bottomrule
	\end{tabular}
	}
\end{table}

\section{Agreement in Relation to Model Performance}
\label{apx:agreement_details}

\begin{figure}
    \centering
    \includegraphics[width=\linewidth]{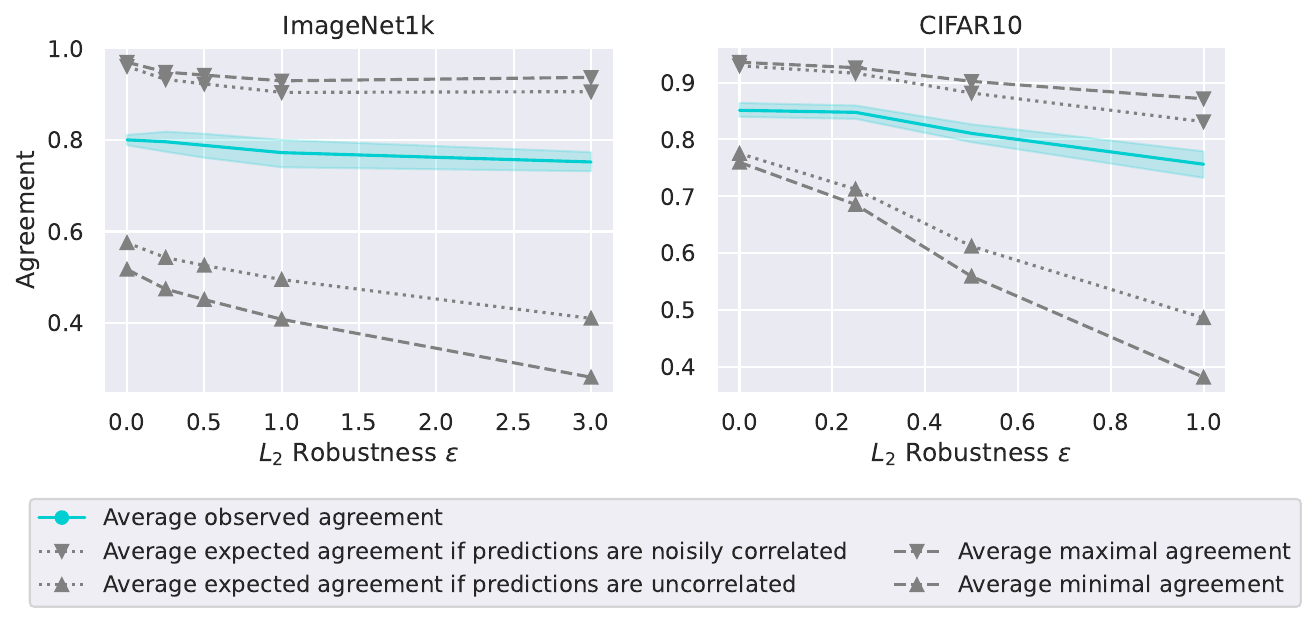}
    \caption{
        \textbf{Lower agreement is not forced by increased differences in model accuracy.}
        While increased robustness leads to larger performance differences between models, which widens the theoretically possible range of agreement, the observed values are not practically limited.
    }
    \label{fig:agreement_details}
\end{figure}

\neu{
In \Cref{subsec:preds_not_universal}, we claim that the decreased agreement of robust models is not explained by larger differences between accuracy.
In \Cref{fig:agreement_details}, we show the average observed agreement, the expected agreement assuming perfectly correlated predictions up to flipping noise and the expected agreement assuming uncorrelated predictions according to \cite{fort2019deep} (dotted lines), as well as theoretical limits to agreement as in \cite{klabunde_prediction_2023} (dashed lines) .
}

\neu{
The observed agreement values are not close to the these limits, which means that higher-than-observed agreement with robust models is theoretically possible.
Relative to the range between minimal and maximal agreement, the observed agreement does increase, however, we note that the scenario for theoretically minimal agreement seems unlikely (correct predictions are overlapping minimally and all instances that are predicted incorrectly by both get different predictions).
}

\section{Jaccard Similarity With Varying Neighborhood Size}
\label{apx:jaccard}

\begin{figure}
    \centering
    \includegraphics[width=\linewidth]{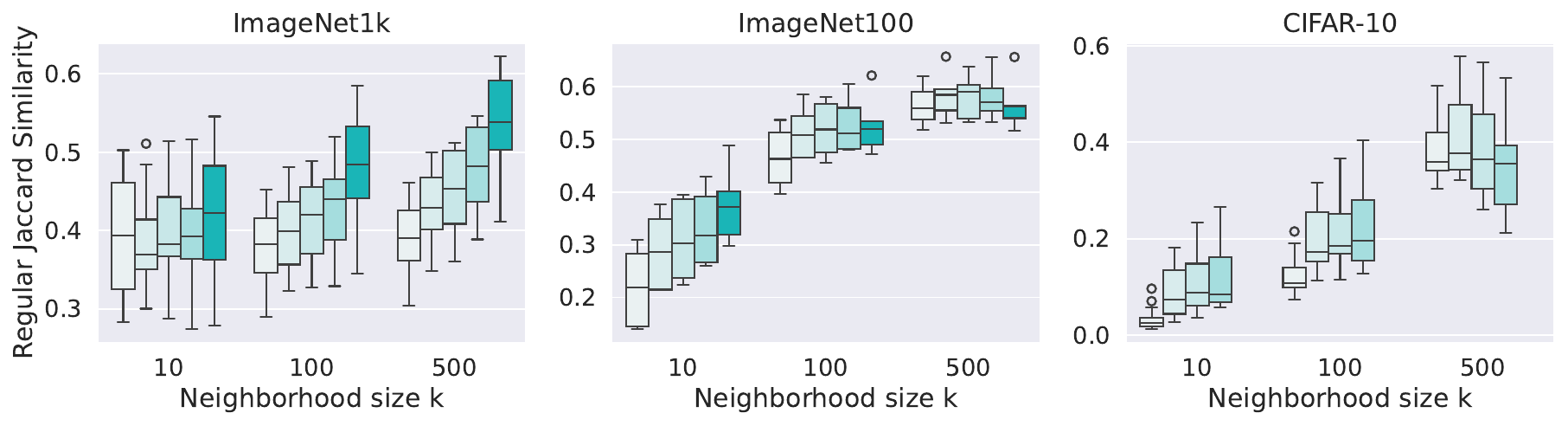}
    \includegraphics[width=\linewidth]{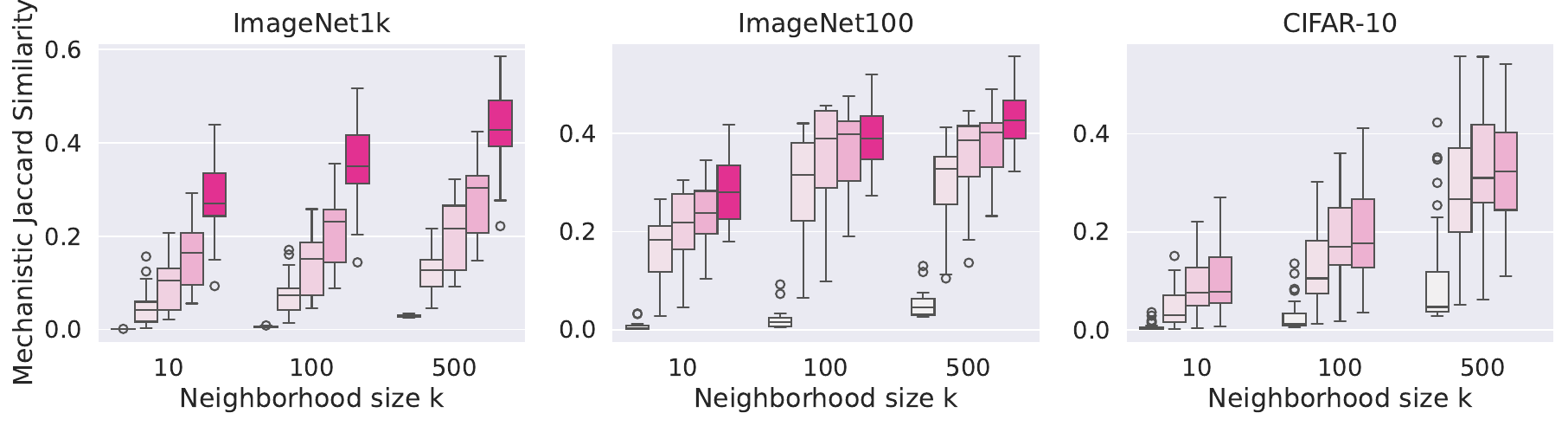}
    \caption{
        \textbf{Jaccard similarity with varying neighborhood size $k$.}
        Neighborhood overlap increases with larger $k$ but trends are similar.
        As $k$ increases, Jaccard Similarity becomes more similar to measures with a global perspective on similarity like CKA.
    }
    \label{fig:jaccard_additional_results}
\end{figure}

\Cref{fig:jaccard_additional_results} shows additional result for Jaccard similarity with neighborhood sizes $k \in \{10, 100, 500\}$.

\section{Compute Resources} \label{apx:compute}
All models were trained using A100s with 80GB memory.
The training time varied depending on the dataset and model size. 
Training on the small CIFAR-10 dataset took around two hours at most using adversarial training. 
Training on ImageNet1k took around one to four days depending on the model.
Execution time for calculating model similarity was likewise dependent on the dataset as well as the measures. 
Reproducing the similarity results shown in this paper would take around 24 hours.

\section{Code and Data} \label{apx:code_and_data}
Our code is available via \url{https://github.com/casparil/rob-univ}. 
Links to the Zenodo repositories for our data are available in our code's README file.

\end{document}